\documentclass[conference]{IEEEtran}
\IEEEoverridecommandlockouts
\usepackage{cite}
\usepackage{amsmath,amssymb,amsfonts}
\usepackage{algorithmic}
\usepackage{graphicx}
\usepackage{textcomp}
\usepackage{xcolor}
\usepackage{booktabs}
\usepackage{multirow}
\usepackage{tabularx}

\usepackage{ragged2e}
\usepackage{adjustbox}
\usepackage[hidelinks]{hyperref}

\usepackage{array}
\newcolumntype{L}[1]{>{\RaggedRight\arraybackslash}p{#1}}
\newcommand{\CDRscan}{CDRscan~\cite{chang2020cdrscan}}
\newcommand{\tCNN}{tCNN~\cite{liu2021tcnn}}
\newcommand{\DeepCDR}{DeepCDR~\cite{kuenzi2020deepcdr}}
\newcommand{\DeepTTA}{DeepTTA~\cite{jiang2022deepTTA}}
\newcommand{\GraTransDRP}{GraTransDRP ~\cite{chu2022graph}}

\def\BibTeX{{\rm B\kern-.05em{\sc i\kern-.025em b}\kern-.08em
    T\kern-.1667em\lower.7ex\hbox{E}\kern-.125emX}}
    
\begin{document}

\title{DeepDTF: Dual-Branch Transformer Fusion for Multi-Omics Anticancer Drug Response Prediction}

\author{

\author{
\IEEEauthorblockN{Yuhan Zhao}
\IEEEauthorblockA{\textit{Department of Computer Science}\\
\textit{North Carolina State University}\\
Raleigh, NC, USA\\
yzhao66@ncsu.edu}
\and
\IEEEauthorblockN{Jacob Tennant}
\IEEEauthorblockA{\textit{Department of Computer Science}\\
\textit{North Carolina State University}\\
Raleigh, NC, USA\\
jmtennan@ncsu.edu}
\and
\IEEEauthorblockN{James Yang}
\IEEEauthorblockA{\textit{Department of Computer Science}\\
\textit{North Carolina State University}\\
Raleigh, NC, USA\\
jbyang2@ncsu.edu}
\and
\IEEEauthorblockN{Zhishan Guo}
\IEEEauthorblockA{\textit{Department of Computer Science}\\
\textit{North Carolina State University}\\
Raleigh, NC, USA\\
zguo32@ncsu.edu}
\and
\IEEEauthorblockN{Young Whang}
\IEEEauthorblockA{\textit{UNC School of Medicine}\\
\textit{University of North Carolina}\\
\textit{at Chapel Hill}\\
Chapel Hill, NC, USA\\
ywhang@med.unc.edu}
\and
\IEEEauthorblockN{Ning Sui\IEEEauthorrefmark{1}}
\IEEEauthorblockA{\textit{Department of Molecular and}\\
\textit{Structural Biochemistry}\\
\textit{North Carolina State University}\\
Raleigh, NC, USA\\
nsui@ncsu.edu}
\thanks{\IEEEauthorrefmark{1}Corresponding author.}
}

}

\maketitle

\begin{abstract}

Cancer drug response varies widely across tumors due to multi-layer molecular heterogeneity, motivating computational decision support for precision oncology. Despite recent progress in deep CDR models, robust alignment between high-dimensional multi-omics and chemically structured drugs remains challenging due to cross-modal misalignment and limited inductive bias. 
We present DeepDTF, an end-to-end dual-branch Transformer fusion framework for joint $\log(\mathrm{IC50})$ regression and drug sensitivity classification. The cell-line branch uses modality-specific encoders for multi-omics profiles with Transformer blocks to capture long-range dependencies, while the drug branch represents compounds as molecular graphs and encodes them with a GNN-Transformer to integrate local topology with global context. Omics and drug representations are fused by a Transformer-based module that models cross-modal interactions and mitigates feature misalignment. On public pharmacogenomic benchmarks under 5-fold cold-start cell-line evaluation, DeepDTF consistently outperforms strong baselines across omics settings, achieving up to \textbf{RMSE $=1.248$}, \textbf{$R^2=0.875$}, and \textbf{AUC $=0.987$} with full multi-omics inputs, while reducing classification error (1-ACC) by 9.5\%.
Beyond accuracy, DeepDTF provides biologically grounded explanations via SHAP-based gene attributions and pathway enrichment with pre-ranked GSEA.
\end{abstract}

\begin{IEEEkeywords}
Drug response prediction, Deep Learning, Graph neural networks, Transformer, Precision oncology, Bioinformatics
\end{IEEEkeywords}

\section{Introduction}

Cancer remains a leading cause of mortality, and treatment outcomes are strongly shaped by genetic factors and the molecular heterogeneity of tumors~\cite{garraway2013precision}. 
Precision oncology aims to tailor therapy to an individual’s molecular profile, making drug response prediction a central computational task~\cite{barretina2012cancer}. 
Because response is influenced by multi-layer molecular mechanisms, effective predictors must model nonlinear interactions between high-dimensional multi-omics features and diverse drug chemical structures~\cite{mcgranahan2017clonal}. 
Compared with empirical screening and in vitro assays, computational modeling offers scalable decision support~\cite{dimasi2016innovation}.

Large pharmacogenomic resources pair molecular profiles with drug sensitivity readouts such as IC50~\cite{sebaugh2011guidelines}, including CCLP~\cite{forbes2017cosmic} and GDSC~\cite{wanjuan2013gdsc}, enabled by advances in high-throughput profiling~\cite{wang2017improved}. 
Earlier methods relied on statistical heuristics and classical machine learning, e.g., Elastic Net with SVM~\cite{wang2016inferences}, random forests~\cite{haider2015copula}, and autoencoders~\cite{xu2019autoencoder}. 
Deep learning reduced reliance on manual feature engineering and improved representation learning, including CNN-based approaches such as CDRscan~\cite{chang2020cdrscan} and tCNN~\cite{liu2021tcnn}, and multimodal designs such as DeepCDR~\cite{kuenzi2020deepcdr} and DeepDSC~\cite{hossein2019deepdsc}. 
However, CNNs can struggle with long-range and context-dependent interactions, motivating Transformer-based modeling~\cite{vaswani2017attention}. 
DeepTTA~\cite{jiang2022deepTTA} integrates SMILES-based drug representations with genomic profiles using Transformer modules, and approaches such as DeepFusionCDR~\cite{hu2024deepfusioncdr} and DeePathNet~\cite{cai2024deepathnet} improve predictive accuracy by modeling cross-modal dependencies and incorporating biological structure. More recent studies further emphasize robust fusion and generalization~\cite{wu2025anticancer}, and transfer-oriented designs such as TransCDR~\cite{xia2024transcdr} that explicitly study generalization to novel compound scaffolds and cell line subgroups.

\begin{figure*}[t]
\vspace{-2mm}
    \centering
    \includegraphics[width=\textwidth]{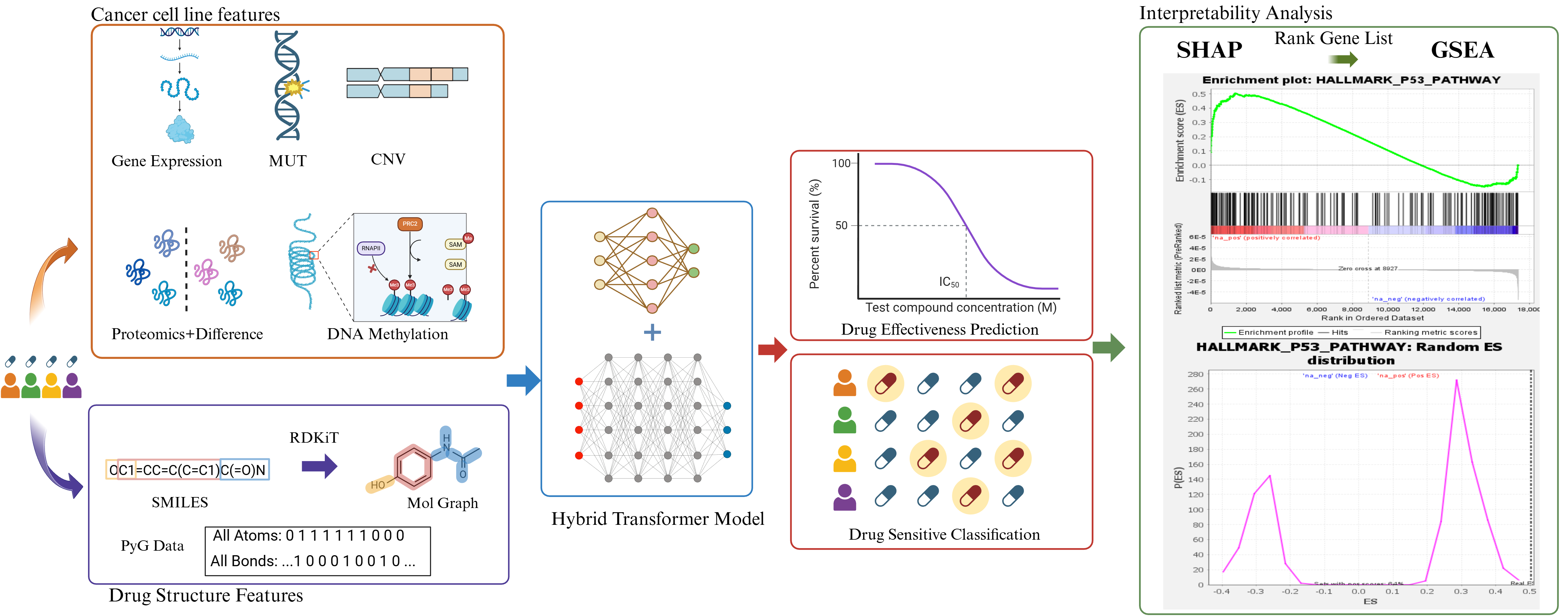}
    
    \caption{Overall workflow of the DeepDTF, including (i) multi-omics and drug-structure inputs, (ii) transformer model (detailed in Fig.~\ref{fig:Detailed_architecture-CNN-transformer}), (iii) dual task prediction (IC50 regression and sensitivity classification), and (iv) an interpretability module based on SHAP$\rightarrow$GSEA. SHAP produces a signed gene ranking, which is used for pre-ranked GSEA: The top panel illustrates a GSEA enrichment plot (running ES with pathway ``hit'' positions along the ranked list), and the bottom panel shows the corresponding null/random ES distribution for assessing enrichment direction and significance.}
    \label{fig:overall_workflow_deepdtf}
    \vspace{-2mm}
\end{figure*}

Effectively modeling interactions between multi-omics features and drug structure remains challenging under high dimensionality and limited sample size. Many pipelines still rely on simplistic fusion strategies such as direct concatenation, static weighting, or shallow attention,  which can under-model cross-modal dependence and semantic misalignment. Moreover, the same drug can exhibit different efficacy under distinct regulatory states, which may not be captured by a single omics layer. This motivates integrating beyond gene-centric signals, such as proteomics and DNA methylation~\cite{mcgranahan2017clonal}. Finally, predictive accuracy alone is insufficient for practical decision support: models benefit from interpretable evidence that can connect predictions to genes and pathways and support hypothesis generation.

Specifically, we propose DeepDTF, a novel hybrid dual-branch Transformer model for anticancer drug response prediction that supports IC50\footnote{The IC50 value, or half-maximal inhibitory concentration, represents the concentration of an inhibitory substance needed to reduce a biological process or response by 50\%. It is a common measure of drug potency, indicating how much of a substance is needed to achieve a specific level of inhibition. A lower IC50 value generally suggests a more potent drug, meaning it requires less substance to achieve the same effect.} regression, sensitivity classification, and interpretability. The cell-line branch integrates heterogeneous omics with a CNN--Transformer to capture local and long-range dependencies, while the drug branch uses a GNN--Transformer to combine molecular-graph topology with global context. A Transformer fusion module then performs dynamic cross-modal attention to align and integrate drug--omics representations, mitigating semantic misalignment. Beyond prediction, we provide an interpretability pipeline based on gene-level SHAP (SHapley Additive exPlanations) ~\cite{lundberg2017shap} attribution and gene set-level pathway enrichment GSEA (Gene Set Enrichment Analysis)~\cite{subramanian2005gsea}, to produce biologically grounded explanations.

\section{Methods}
\label{sec2methods}
\subsection{Overview of DeepDTF Framework}

DeepDTF targets core challenges in cancer drug response (CDR) modeling: learning reliable drug--cell-line interactions under high-dimensional, heterogeneous omics and chemically structured drugs, where naive fusion can suffer from cross-modal misalignment.
To address this, we adopt a dual-branch design (Fig.~\ref{fig:overall_workflow_deepdtf}) that first extracts modality-preserving token representations for (i) cell-line multi-omics and (ii) drug structure, and then performs token-level fusion to model their interactions for $\log(\mathrm{IC50})$ regression and sensitivity classification.
Beyond predictive accuracy, DeepDTF provides biologically grounded explanations by connecting model attributions to genes and pathways through a post-hoc interpretability pipeline based on gene-level SHAP and pathway-level GSEA. Details of the interpretability workflow, experimental protocol, and biological findings are presented in Sec.~\ref{sec:interpretability_section}.

\begin{figure*}[t]
    \centering
    \includegraphics[width=\linewidth]{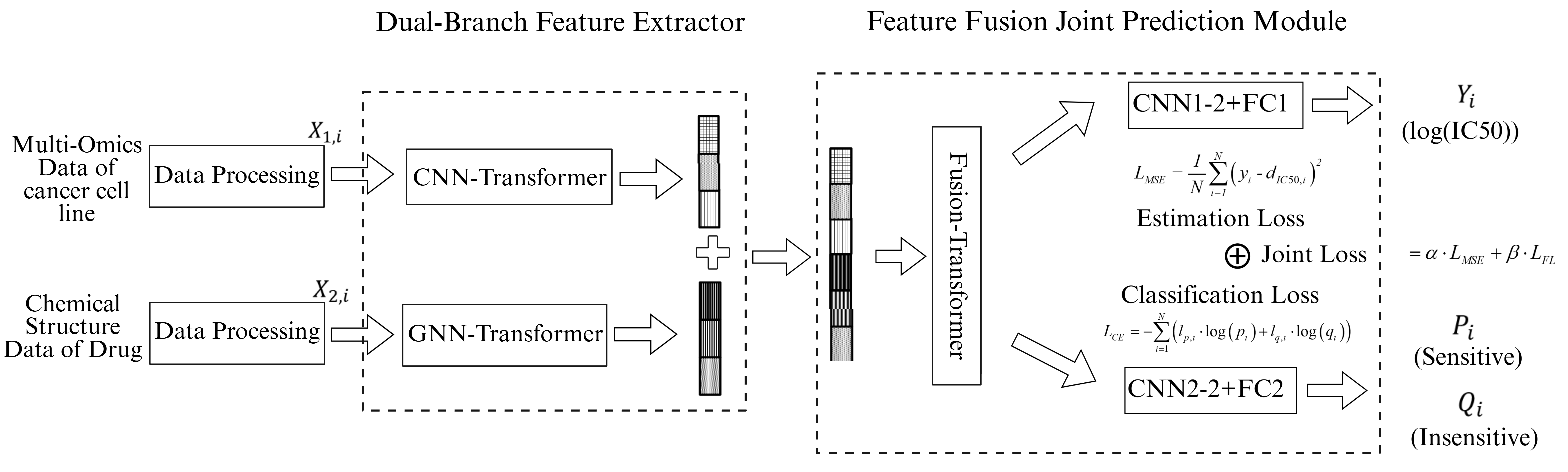}
    \caption{Detailed architecture of DeepDTF. The left block is a dual-branch feature extractor for multi-omics and drug structure. The right block is a feature fusion joint prediction module with a Fusion-Transformer and two task-specific heads for IC50 regression and sensitivity classification.}
    \label{fig:Detailed_architecture-CNN-transformer}

\end{figure*}

Figure~\ref{fig:Detailed_architecture-CNN-transformer} details the architecture. 
In order to capture long-range dependencies among heterogeneous omics signals, the cell-line branch uses modality-specific encoders followed by Transformer blocks. 
In order to preserve chemical topology and functional substructures, the drug branch encodes molecular graphs with a GNN-Transformer.
Finally, to mitigate semantic misalignment between omics and drug features, a Fusion-Transformer performs cross-modal attention over the joint token set, followed by lightweight task heads for regression and classification.

\subsection{Input Representation and Preprocessing}
\label{sec:input_repr}

\noindent\textbf{Multi-omics inputs.} 
Multi-omics signals are heterogeneous. To preserve modality semantics while enabling joint modeling, we maintain separate modality-aligned vectors and denote the feature set for each cell line $c$ as
\begin{equation}
\mathcal{X}_c = \left\{ \mathbf{x}^{\mathrm{GE}}_c, \mathbf{x}^{\mathrm{MUT}}_c, \mathbf{x}^{\mathrm{CNV}}_c, \mathbf{x}^{\mathrm{PROT}}_c, \mathbf{x}^{\Delta\mathrm{PROT}}_c, \mathbf{x}^{\mathrm{METH}}_c \right\},
\end{equation}
where GE/MUT/CNV denote gene expression / mutation / copy-number variation, $\mathrm{PROT}$ denotes proteomics abundance, $\Delta\mathrm{PROT}$ denotes proteomics difference with respect to healthy references, and $\mathrm{METH}$ denotes DNA methylation features.

\noindent\textbf{Drug graph inputs.}
Preserving local chemical context and topology is critical for accurate and explainable drug response modeling, since many drug effects are driven by specific functional groups and local substructures. 
We convert each drug $d$ from its SMILES string into a molecular graph $G_d=(V_d,E_d)$ using RDKit~\cite{landrum2013rdkit} and represent it as a PyTorch Geometric (PyG) object.
Each atom $v\in V_d$ is assigned a feature vector $\mathbf{h}_v$, and each bond $(u,v)\in E_d$ is assigned an edge feature vector $\mathbf{e}_{uv}$.

\vspace{-1mm}
\subsection{Dual-Branch Feature Extractor}
\label{sec:dual_branch}

Multi-omics profiles are high-dimensional feature vectors without a universally agreed interaction graph, making graph-based inductive biases unreliable. 
In order to obtain compact tokens while retaining local feature patterns, we therefore tokenize multi-omics features via CNN-based projection. 
In contrast, drug molecules have a chemically grounded graph structure, where message passing captures local chemical context and functional groups; thus we preserve this inductive bias via GNN-based encoding for drug structures (Fig.~\ref{fig:Detailed_architecture-CNN-transformer} left).

\subsubsection{Cell-line Branch: CNN with Attention + Transformer}
\label{sec:cell_branch}

\textbf{CNN tokenization with attention.} We first apply a CNN-based encoder with attention to capture local patterns and produce a compact token sequence. 
Let $\mathbf{x}_c$ denote the modality-stacked omics feature vector, then the CNN encoder outputs a sequence of latent tokens:
\begin{equation}
\mathbf{H}^{(0)}_c = f_{\mathrm{CNN\text{-}Attn}}(\mathbf{x}_c), \qquad \mathbf{H}^{(0)}_c \in \mathbb{R}^{n_c \times d}.
\end{equation}
The channel attention module uses multi-scale convolutions and global pooling to reweight feature channels, improving robustness under high dimensionality and limited samples.

\noindent\textbf{Transformer encoding.} We then apply a Transformer encoder to $\mathbf{H}^{(0)}_c$ to capture long-range dependencies among omics-derived tokens:
\begin{equation}
\mathbf{H}_c = \mathrm{Trans}_{\mathrm{omics}}(\mathbf{H}^{(0)}_c).
\end{equation}
Each Transformer layer uses multi-head self-attention. The output is followed by residual connections, layer normalization, and a position-wise feed-forward network (FFN), as in standard Transformer designs.

\subsubsection{Drug Branch: GNN + Transformer}
\label{sec:drug_branch}

\noindent\textbf{Graph neural encoding for local chemical context.}
Given a molecular graph $G_d=(V_d,E_d)$, we apply a graph neural network (GNN) to obtain node embeddings. 
A generic message-passing layer can be written as:
\begin{equation}
\mathbf{m}_u^{(\ell)}=\sum_{v\in\mathcal{N}(u)} \phi^{(\ell)}\!\left(\mathbf{h}_u^{(\ell)},\mathbf{h}_v^{(\ell)},\mathbf{e}_{uv}\right), \end{equation}
\begin{equation}
\mathbf{h}_u^{(\ell+1)} = \psi^{(\ell)}\!\left(\mathbf{h}_u^{(\ell)},\mathbf{m}_u^{(\ell)}\right),
\end{equation}
where $\mathcal{N}(u)$ denotes the neighbors of node $u$, and $\phi,\psi$ are learnable functions, like MLPs with normalization. 
After $L$ layers, we obtain node token embeddings:
\begin{equation}
\mathbf{H}^{(0)}_d = [\mathbf{h}_u^{(L)}]_{u\in V_d} \in \mathbb{R}^{n_d \times d}.
\end{equation}

\noindent\textbf{Transformer encoding.} While GNNs capture local topology effectively, long-range dependencies and global substructure interactions can be challenging for shallow message passing. 
We therefore apply a Transformer encoder over the node tokens:
\begin{equation}
\mathbf{H}_d = \mathrm{Trans}_{\mathrm{drug}}(\mathbf{H}^{(0)}_d),
\end{equation}
so that the drug representation benefits from both GNN and global token interactions.

\vspace{-1mm}
\subsection{Feature Fusion Joint Prediction Module}
\label{sec:fusion_prediction}

After obtaining omics tokens $\mathbf{H}_c$ and drug node tokens $\mathbf{H}_d$, DeepDTF performs explicit cross-modal interaction modeling using a Fusion-Transformer (Fig.~\ref{fig:Detailed_architecture-CNN-transformer} right). 
We concatenate the two token sequences and feed them into a Fusion-Transformer:
\begin{equation}
\mathbf{H}_{\mathrm{joint}} = \mathrm{Trans}_{\mathrm{fusion}}([\mathbf{H}_c;\mathbf{H}_d]).
\end{equation}
Because self-attention is computed over the joint token set, $\mathrm{Trans}_{\mathrm{fusion}}$ can dynamically align drug substructures with context-dependent cellular states, mitigating semantic misalignment between modalities.

A learnable pooling operator produces a pair-level embedding $\mathbf{z}$: 
\begin{equation}
\mathbf{z} = \mathrm{Pool}(\mathbf{H}_{\mathrm{joint}})\in\mathbb{R}^{d}.
\end{equation}

Then pooled representation $\mathbf{z}$ is passed to two prediction heads:
(i) a regression head to predict $\log(\mathrm{IC50})$,
and (ii) a classification head to predict sensitivity:
\begin{equation}
\hat{y} = g_{\mathrm{reg}}(\mathbf{z}), \qquad \hat{p} = g_{\mathrm{cls}}(\mathbf{z}).
\end{equation}

We optimize a multi-task objective with MSE for regression and focal loss for classification:
\begin{equation}
\mathcal{L}_{\mathrm{reg}} = \frac{1}{N}\sum_{i=1}^{N}(\hat{y}_i - y_i)^2,
\end{equation}
\begin{equation}
\mathcal{L}_{\mathrm{FL}} = -\frac{1}{N}\sum_{i=1}^{N}\Big[ (1-\hat{p}_i)^\gamma t_i\log(\hat{p}_i) + \hat{p}_i^\gamma(1-t_i)\log(1-\hat{p}_i)\Big],
\end{equation}
where $t_i\in\{0,1\}$ is the ground-truth sensitivity label and $\gamma>0$ controls the focusing strength.

The overall objective is a weighted multi-task loss with $\ell_2$ regularization:
\begin{equation}
\mathcal{L} = \alpha\,\mathcal{L}_{\mathrm{reg}} + \beta\,\mathcal{L}_{\mathrm{FL}} + \lambda \|\Theta\|_2^2,
\end{equation}
where $\Theta$ denotes trainable parameters and $\alpha,\beta,\lambda$ are hyperparameters.

\section{Experiments and results}
\subsection{Dataset and Pre-Processing}
\label{subsec:data_preproc}

\paragraph{Data sources}
We use GDSC2 (Release 8.5, Oct 2023)~\cite{wanjuan2013gdsc} for IC50 and genomics, CCLP~\cite{forbes2017cosmic} for multi-omics, and PubChem~\cite{kim2016pubchem} for standardized drug structures (SMILES) and identifiers. To ensure reliable training and sufficient representation per category, we retain cancer types with at least 10 cell lines.
We further keep cell lines whose mutation/CNV profiles overlap with Cancer Gene Census (CGC) genes from CCLP to improve feature reliability.
For drugs, we exclude compounds without valid PubChem compound IDs and remove molecules with molecular weight $>1000$ g/mol to retain drug-like candidates. This resulted in 782 cell lines in 28 cancer types and 256 anticancer drugs.

\paragraph{Data preprocessing}

\hfill\break
\hspace*{1 em} \textit{Genomic:} We normalize gene expression (GE) profiles from GDSC2 using the Robust Multi-array Average (RMA) algorithm to [0,1] yielding 17{,}777 gene features per cell line.
Mutation (MUT) data are encoded as binary vectors of length 28,669 (1 for mutation, 0 otherwise), and CNV data as ternary vectors of length 735 ($-1$ for deletion, $0$ for normal, $+1$ for amplification).

\textit{Proteomics:} We incorporate proteomics abundance with 8{,}453 proteins per cell line. 
A consistent subset of 822 proteins is observed across all cell lines, while the remaining proteins exhibit varying missingness.
We additionally derive a proteomics-difference view ($\Delta$PROT) by contrasting each cell line against a healthy baseline from ProteomicsDB~\cite{proteomicsDB}, i.e., $\mathbf{x}^{\Delta\mathrm{PROT}}_c=\mathbf{x}^{\mathrm{PROT}}_c-\mathbf{x}^{\mathrm{PROT}}_{\mathrm{healthy}}$.
We use Additive Shift Weighting (ASW) as the default integration:
\begin{equation}
\tilde{\mathbf{x}}^{\mathrm{PROT}}_c = \mathbf{x}^{\mathrm{PROT}}_c + \mathbf{x}^{\Delta\mathrm{PROT}}_c.
\end{equation}

\textit{DNA methylation:} We include DNA methylation features represented by 56{,}146 methylation clusters per cell line, where each cluster contains 846 attributes including genomic coordinates and coverage-related quality indicators. 
Methylation values are in $[0,1]$, coverage is defined as (total read depth of a gene fragment) / (number of CpGs in the cluster), and is used for quality control. We used coverage $\ge 10$ to remove low-confidence clusters while retaining most loci, following common practice in bisulfite-seq analysis~\cite{akalin2012methylkit}.

\textit{Drug structure graphs:}
We represent each drug as a 2D molecular graph constructed from its SMILES string using RDKit~\cite{landrum2013rdkit} and processed with PyTorch Geometric.
Following the widely used OGB featurization, each atom is associated with a 9-dimensional categorical node feature and each bond is associated with a 3-dimensional categorical edge feature. 

After filtering, the final dataset contains 164{,}165 drug--cell line pairs, spanning 782 cell lines and 256 drugs across 28 cancer types.
For classification, we binarize responses using the standard threshold~\cite{iorio2016landscape}:
\begin{equation}
\mathrm{Label}_{d,c}=
\begin{cases}
1, & \log(\mathrm{IC50})_{d,c} < -2.0 \quad (\text{Sensitive}) \\
0, & \text{otherwise} \quad (\text{Resistant})
\end{cases}
\end{equation}

\vspace{-1.5mm}
\subsection{Evaluation Metrics}
\vspace{-1mm}
We report standard metrics for both tasks, chosen to quantify different aspects of prediction quality in CDR modeling.

\paragraph{Regression (IC50)}
\begin{itemize}
\item \textbf{RMSE:} the typical magnitude of prediction error (in log-IC50 units). Lower RMSE means predictions are, on average, closer to ground truth.
\item \textbf{$R^2$:} the fraction of variance in the true IC50 values explained by the model. Higher $R^2$ indicates better fit to the response distribution.
\item \textbf{PCC:} the strength of linear association between predicted and true IC50 values. Higher PCC indicates better ranking consistency and trend alignment.
\end{itemize}

\paragraph{Classification (Sensitivity)}
\begin{itemize}
\item \textbf{AUC:} measure how well the model ranks sensitive vs.\ resistant pairs; robust under class imbalance.
\item \textbf{ACC:} overall fraction of correctly predicted labels.
\item \textbf{SEN (Recall for Sensitive):} the ability to correctly identify truly sensitive cases (avoid missing effective drugs).
\item \textbf{SPEC (Recall for Resistant):} the ability to correctly identify truly resistant cases (avoid recommending ineffective drugs).
\end{itemize}

\begin{table*}[t]
\centering
\small
\setlength{\tabcolsep}{10.0pt}
\renewcommand{\arraystretch}{0.95}
\caption{Evaluation of the Proposed Method and Baselines}
\label{tab:merged_results}

\begin{tabularx}{\textwidth}{@{} L{0.09\textwidth} L{0.16\textwidth} c c c c c c c @{}}
\toprule
Setting & Method & RMSE & $R^2$ & PCC & ACC & SEN & SPEC & AUC \\
\midrule

\multirow{4}{*}{MUT+CNV} 
& \tCNN  & 1.459 & 0.725 & 0.851 & 0.952 & 0.706 & 0.977 & 0.963 \\
& \DeepCDR  & \textbf{1.401} & 0.770 & 0.878 & 0.953 & 0.732 & 0.975 & 0.969 \\
& \DeepTTA  & 1.420 & 0.775 & 0.880 & 0.953 & 0.732 & \textbf{0.978} & 0.970 \\
& DeepDTF   & 1.403 & \textbf{0.780} & \textbf{0.883} & \textbf{0.954} & \textbf{0.738} & \textbf{0.978} & \textbf{0.975} \\
\midrule

\multirow{5}{*}{MUT+CNV+GE} 
& \CDRscan  & 1.483 & 0.715 & 0.846 & 0.950 & 0.667 & 0.979 & 0.960 \\
& \tCNN     & 1.459 & 0.725 & 0.851 & 0.952 & 0.706 & 0.977 & 0.963 \\
& \DeepCDR  & 1.320 & 0.822 & 0.907 & 0.958 & 0.754 & 0.975 & 0.975 \\
& \DeepTTA  & 1.302 & 0.821 & 0.906 & 0.958 & 0.760 & 0.980 & 0.976 \\
& DeepDTF   & \textbf{1.253} & \textbf{0.870} & \textbf{0.933} & \textbf{0.962} & \textbf{0.768} & \textbf{0.981} & \textbf{0.982} \\
\midrule
\multirow{5}{*}{All Multi-Omics} 
& \CDRscan  & 1.480 & 0.717 & 0.846 & 0.951 & 0.668 & 0.981 & 0.961 \\
& \tCNN     & 1.458 & 0.725 & 0.852 & 0.954 & 0.708 & 0.980 & 0.965 \\
& \DeepCDR  & 1.312 & 0.824 & 0.910 & 0.959 & 0.754 & 0.976 & 0.975 \\
& \DeepTTA  & 1.292 & 0.825 & 0.912 & 0.959 & 0.756 & 0.980 & 0.977 \\
& \GraTransDRP  & 1.272 & 0.850 & 0.922 & 0.964 & 0.732 & 0.981 & 0.980 \\
& DeepDTF   & \textbf{1.248} & \textbf{0.875} & \textbf{0.938} & \textbf{0.965} & \textbf{0.771} & \textbf{0.983} & \textbf{0.987} \\
\midrule
\bottomrule
\end{tabularx}
\vspace{-2mm}
\end{table*}

\vspace{-1.5mm}
\subsection{Baselines}
We compare DeepDTF with representative deep CDR predictors under identical splits and preprocessing.

\textbf{CDRscan}~\cite{chang2020cdrscan} encodes mutation features and drug fingerprints via CNN modules and integrates them with a docking-inspired interaction mechanism for IC50 prediction.

\textbf{tCNN}~\cite{liu2021tcnn} uses dual CNN encoders for drug and cell line features (mutation/CNV) followed by feature fusion for response prediction.

\textbf{DeepCDR}~\cite{kuenzi2020deepcdr} is a strong multimodal framework combining drug molecular graphs with multi-omics (e.g., GE/MUT/CNV) using graph convolution and omics-specific subnetworks.

\textbf{DeepTTA}~\cite{jiang2022deepTTA} adopts Transformer-based drug representation learning (from SMILES tokens) with an MLP for transcriptomic features, supporting both IC50 regression and sensitivity classification.

\textbf{GraTransDRP} ~\cite{chu2022graph} introduces a graph-transformer-style architecture for drug response prediction, and public implementations support multi-omics inputs (methylation + gene expression + mutation) together with drug graphs.

\vspace{-1.5mm}
\subsection{Experimental Results and Comparison Study}

We implement DeepDTF in PyTorch\footnote{Code will be released upon acceptance.} and train with Adam for 60 epochs (batch size 256, learning rate $1\times10^{-3}$, weight decay $3\times10^{-4}$; focal loss $\gamma=2$). 
We select checkpoints by validation RMSE within each fold and run experiments on an NVIDIA RTX 3090 GPU.

Cold start cell line evaluation:
We use 5-fold cross-validation with cell-line-level splits stratified by cancer type.
All drug-cell line pairs inherit the fold assignment of their cell line, so the test folds contain unseen cell lines. This prevents cell-line leakage through repeated pairing of the same cell line with multiple drugs.
Feature scaling is fit on the training split of each fold and applied to the corresponding test split. We keep the drug set fixed and evaluate generalization to new cell lines, matching drug repositioning and screening settings.

Table~\ref{tab:merged_results} summarizes the comparison under three input settings and the results are averaged over 5 folds. 
For \textbf{MUT+CNV}, DeepDTF achieves $R^2=0.780$ and AUC$=0.975$, yielding the best overall.
When adding \textbf{GE} (MUT+CNV+GE), DeepDTF outperforms all competitors on both regression and classification: it reduces RMSE by $\sim$3.8\% compared with the best baseline and increases AUC from 0.976 to 0.982; meanwhile, the classification error rate (1-ACC) decreases by 9.5\%. 
With \textbf{All Multi-Omics} enabled, DeepDTF further improves AUC to $=0.987$, surpassing GraTransDRP across all metrics, indicating that richer omics signals and explicit cross-modal token fusion jointly contribute to more accurate and reliable CDR prediction.

\vspace{-2mm}
\begin{figure}[htbp]
    \centering
    \includegraphics[width=\linewidth]{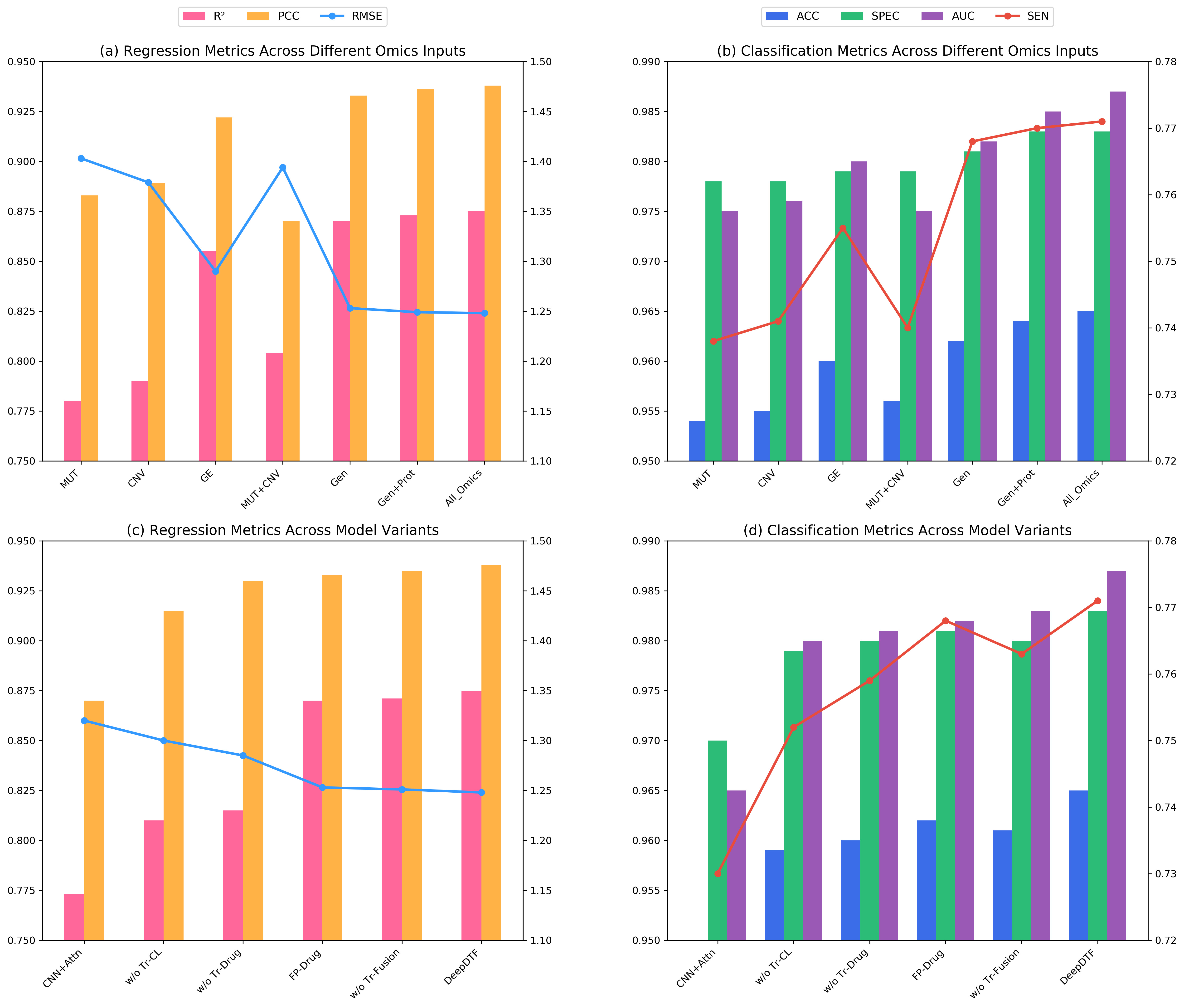}
    \vspace{-5mm}
    \caption{Ablation studies on omics integration and model components, where (a) and (c) show the regression performance under different combinations of omics inputs and model variants; while (b) and (d) report the classification metrics obtained from various omics integration strategies and model variants.}
    \label{fig:ablation_study}
    \vspace{-2mm}
\end{figure}

\vspace{-1.5mm}
\subsection{Ablation Studies}
\vspace{-1.5mm}

We conduct two ablation studies to examine (i) the effect of omics inputs and (ii) the contribution of key model modules, under identical training and evaluation protocols.

\textbf{Omics ablation.}
As shown in Fig.~\ref{fig:ablation_study}(a), regression performance under different omics inputs is visualized using RMSE (left Y-axis), $R^2$, and PCC (right Y-axis). Classification results are shown in Fig.~\ref{fig:ablation_study}(b) via ACC, SEN, SPEC, and AUC. 
GE is the strongest single modality, and adding modalities consistently improves performance.
Proteomics further improves performance, and All\_Omics yields the best overall results (RMSE=1.248, $R^2$=0.875, PCC=0.938, AUC=0.987), indicating complementary gains from additional omics layers.

\textbf{Model ablation.}
Fig.~\ref{fig:ablation_study}(c,d) evaluate six variants: CNN+Attn, w/o Tr-CL, w/o Tr-Drug, FP-Drug, w/o Tr-Fusion, and DeepDTF.
Removing either the cell-line Transformer (w/o Tr-CL) or the drug Transformer (w/o Tr-Drug) degrades both regression and classification, confirming the importance of token-level global modeling on both modalities. 
Replacing drug graphs with fingerprints (FP-Drug) is consistently worse than DeepDTF, supporting the benefit of graph-based drug structure modeling.
Finally, removing the Fusion-Transformer slightly reduces performance, indicating additional gains from explicit cross-modal interaction modeling.

\section{Interpretability Analysis}
\label{sec:interpretability_section}
High predictive accuracy is necessary but insufficient for precision oncology, where mechanistic evidence is often required to trust a model and to generate hypotheses. We therefore perform an interpretability analysis that links DeepDTF predictions to gene and pathway-level signals (Fig.~\ref{fig:overall_workflow_deepdtf}).

We first use SHAP~\cite{lundberg2017shap} on the drug sensitivity classification output to obtain signed gene attributions, indicating which genes drive higher or lower IC50. SHAP gives direction-aware gene contributions, but single gene lists are unstable and hard to interpret mechanistically. We therefore project the signed SHAP ranking onto curated gene sets using pre-ranked GSEA~\cite{subramanian2005gsea} with MSigDB Hallmark~\cite{liberzon2015hallmark} to summarize genes into pathway programs.
Together, SHAP identifies direction-aware gene drivers, while GSEA maps these drivers to interpretable biological mechanisms, yielding biologically grounded explanations at both gene and pathway levels.

\subsection{Protocol}
SHAP assigns each gene-aligned input features $j$ a signed contribution score $\phi_j$ for a given prediction, so that
\begin{equation}
\hat{y} \approx \phi_0 + \sum_{j=1}^{M}\phi_j,
\end{equation}
where $\phi_0$ is the expected output under a background distribution.
We adopt the signed convention:
\begin{itemize}
\item $\phi_j>0$: gene $j$ increases the predicted sensitivity score (sensitivity-associated contribution).
\item $\phi_j<0$: gene $j$ decreases the predicted sensitivity score (resistance-associated contribution).
\end{itemize}
Thus, SHAP provides both importance and direction for mechanistic interpretation.

We then perform pre-ranked GSEA on MSigDB Hallmark gene sets to obtain pathway-level enrichments~\cite{subramanian2005gsea,liberzon2015hallmark}. We rank genes by aggregated signed SHAP scores and compute an enrichment score (ES) with direction:
\begin{itemize}
\item ES $>0$: the pathway is enriched among sensitivity-driving genes (positive SHAP).
\item ES $<0$: the pathway is enriched among resistance-driving genes (negative SHAP).
\end{itemize}
In this way, the SHAP $\rightarrow$ GSEA procedure converts local feature attributions into statistically supported pathway programs, making the explanations more robust and biologically interpretable.

\vspace{-2mm}
\begin{figure}[t]
    \centering
    \includegraphics[width=\linewidth]{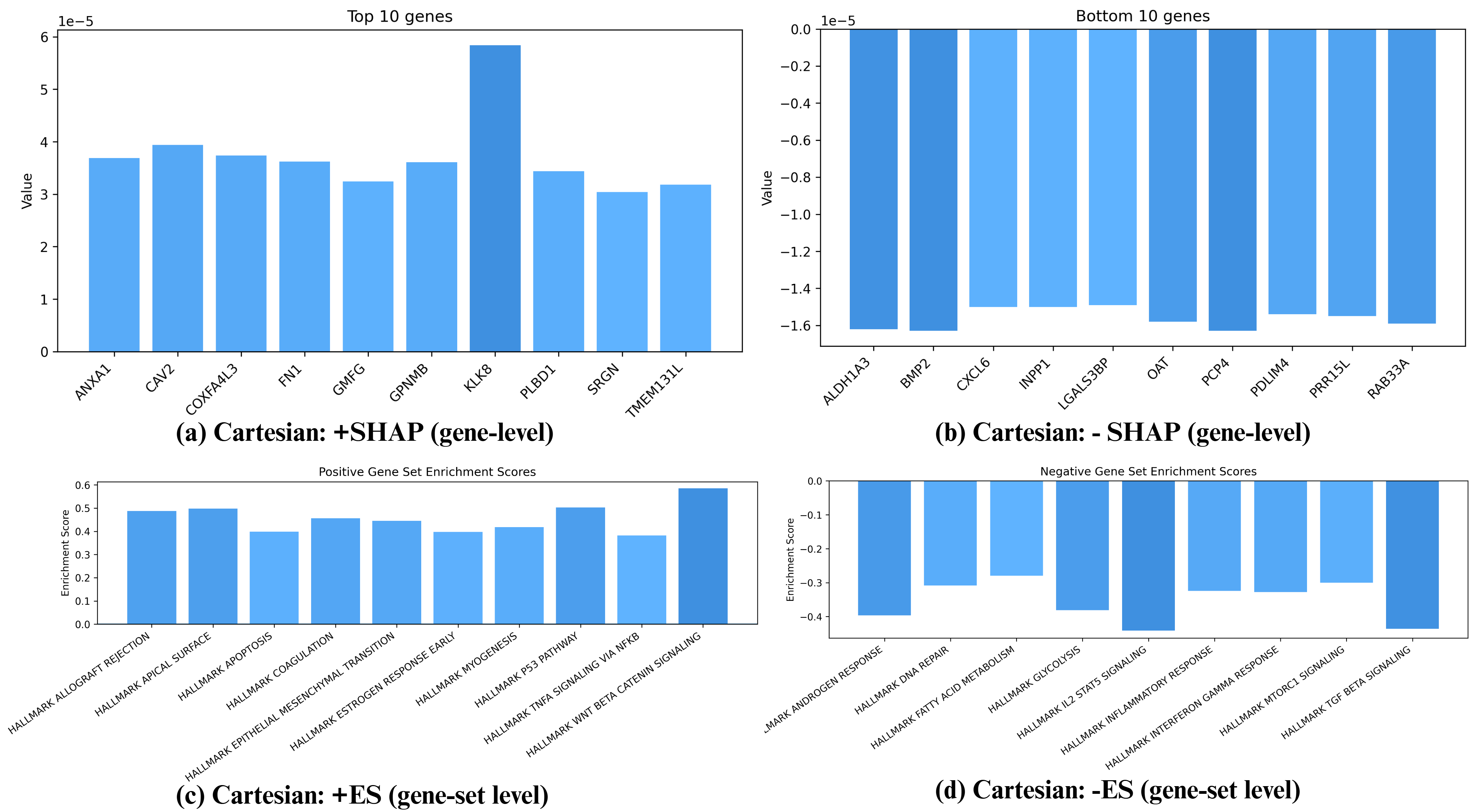}
    \vspace{-5mm} 
    \caption{Cartesian-regime biological explanations via SHAP--GSEA. (a) Top genes with positive SHAP (sensitivity-associated). (b) Bottom genes with negative SHAP (resistance-associated). (c) Top enriched Hallmark pathways from signed-SHAP rankings (positive ES; sensitivity-associated programs). (d) Top enriched Hallmark pathways from negative signed-SHAP rankings (negative ES; resistance-associated programs)}
    \label{fig:shap_gsea_cartesian}
    \vspace{-3mm}
\end{figure}

\subsection{Results and interpretation}
\vspace{-1.5mm}

We report results under a cancer$\times$drug Cartesian evaluation regime: for each cancer type, we aggregate SHAP values within each cancer type across a broad set of drug contexts, aiming to capture cancer-contextual signals.

Figs.~\ref{fig:shap_gsea_cartesian}(a) and (b) report the top genes with the largest positive and negative SHAP values, respectively, highlighting the most influential sensitivity- and resistance-associated gene drivers.
Figs.~\ref{fig:shap_gsea_cartesian}(c) and (d) report Hallmark pathways enriched from the signed-SHAP rankings. We observe that several well-established cancer-related pathways are significantly enriched among the highest-attribution genes.
On the sensitivity-associated side (Fig.~\ref{fig:shap_gsea_cartesian}c), enriched pathways such as \textit{TNFA\_SIGNALING\_VIA\_NFKB}, \textit{WNT\_BETA\_CATENIN\_SIGNALING}, and \textit{P53\_PATHWAY} indicate that the model relies on coordinated inflammatory, stress-response and oncogenic-state programs when classifying a drug as sensitive.
On the resistance-associated side (Fig.~\ref{fig:shap_gsea_cartesian}d), enriched pathways such as \textit{DNA\_REPAIR}, \textit{MTORC1\_SIGNALING}, and \textit{INTERFERON\_GAMMA\_RESPONSE} suggest that DeepDTF leverages coordinated programs related to genome maintenance, growth/metabolism regulation, and immune-related signaling when classifying a drug as resistant.

Overall, the SHAP$\rightarrow$GSEA analysis provides more than post-hoc visualization: it offers a mechanism-aware sanity check and a hypothesis-generation tool.
DeepDTF consistently re-identifies multiple well-characterized cancer programs as the dominant decision drivers, offering strong face validity that the model’s decisions are anchored in coherent biology rather than unstructured feature noise. 
Importantly, beyond recovering established pathways, the attribution rankings also surface less-emphasized genes and pathway programs that may have been overlooked in standard biomarker-centric analyses, thereby generating testable hypotheses for follow-up experiments. 
Such signals can prioritize follow-up studies (e.g., targeted validation, biomarker stratification, and resistance mechanism probing) and may ultimately support precision oncology workflows, including mechanism-guided patient stratification or personalized drug design by highlighting which molecular programs should be perturbed to shift a predicted response.

\section{Conclusion}
\label{sec:conclusion}
We presented DeepDTF, an end-to-end dual-branch framework for cancer drug response prediction that jointly models cancer cell line multi-omics and drug chemical structure. DeepDTF uses a CNN--Transformer encoder to tokenize and contextualize heterogeneous omics signals and a GNN-Transformer encoder to capture both local chemical topology and global substructure dependencies from molecular graphs. A Fusion-Transformer then performs fine-grained cross-modal attention over omics and drug tokens, enabling context-dependent interaction modeling and mitigating semantic misalignment. Across multiple omics settings, DeepDTF consistently improves both $\log(\mathrm{IC50})$ regression and sensitivity classification over strong baselines, and ablation studies verify the contributions of omics augmentation, graph-based drug encoding, and each Transformer component. To enhance practical transparency, we further provide a post-hoc interpretability pipeline that links gene-level SHAP attributions to pathway-level programs via pre-ranked GSEA, yielding direction-aware, biologically coherent explanations. Together, these results suggest DeepDTF as a strong and interpretable foundation for precision oncology decision support.

\bibliographystyle{IEEEtran}
\bibliography{references}

\end{document}